# A Vessel Bifurcation Landmark Pair Dataset for Abdominal CT Deformable Image Registration (DIR) Validation

**Authors**: Edward R Criscuolo[1], Yao Hao[2], Zhendong Zhang[1], Trevor McKeown[1], Deshan Yang[1]

**Author Affiliations**

1. Department of Radiation Oncology, Duke University, Durham, NC, 27701, USA

2. Washington University School of Medicine, St. Louis, MO, 63110, USA

**Corresponding Author**

Deshan Yang, PhD, DABR

Professor, Department of Radiation Oncology, School of Medicine, Duke University

deshan.yang@duke.edu

40 Duke Medicine Circle, Room 04212

3640 DUMC

Durham, NC 27710

# Abstract

**Purpose**

Deformable image registration (DIR) is an enabling technology in many diagnostic and therapeutic tasks. Despite this, DIR algorithms have limited clinical use, largely due to a lack of benchmark datasets for quality assurance during development. DIRs of intra-patient abdominal CTs are among the most challenging registration scenarios due to significant organ deformations and inconsistent image content. To support future algorithm development, here we introduce our first-of-its-kind abdominal CT DIR benchmark dataset, comprising large numbers of highly accurate landmark pairs on matching blood vessel bifurcations.

**Acquisition and Validation Methods**

Abdominal CT image pairs of 30 patients were acquired from several publicly available repositories as well as the authors' institution with IRB approval. The two CTs of each pair were originally acquired for the same patient but on different days. An image processing workflow was developed and applied to each CT image pair: 1) Abdominal organs were segmented with a deep learning model, and image intensity within organ masks was overwritten. 2) Matching image patches were manually identified between two CTs of each image pair. 3) Vessel bifurcation landmarks were labeled on one image of each image patch pair. 4) Image patches were deformably registered, and landmarks were projected onto the second image 5) Landmark pair locations were refined manually or with an automated process. This workflow resulted in 1895 total landmark pairs, or 63 per case on average. Estimates of the landmark pair accuracy using digital phantoms were 0.7mm +/- 1.2 mm.

**Data Format and Usage Notes**

The data is published in Zenodo at https://doi.org/10.5281/zenodo.14362785. Instructions for use can be found at https://github.com/deshanyang/Abdominal-DIR-QA.

**Potential Applications**

This dataset is a first-of-its-kind for abdominal DIR validation. The number, accuracy, and distribution of landmark pairs will allow for robust validation of DIR algorithms with precision beyond what is currently available.

# 1 Introduction

Image registration and fusion algorithms have become prevalent in modern medicine with the increased integration of imaging into diagnostic and therapeutic pipelines[1,2]. Providers must often consider several images, potentially acquired at different times and via different imaging modalities, jointly to effectively diagnose or plan treatment for a patient[3]. Deformable image registration (DIR) is one tool available for such tasks and involves finding the voxel-wise geometric transformation to align anatomies in two images[4]. This transformation is modeled by a deformation vector field (DVF), which maps a moving image onto the target image. Unlike rigid registration, where voxels move uniformly and voxel-voxel relationships are maintained, voxels in DIR can move independently of one another, which can help model complex motion and deformations found in soft tissue[5]. DIR has many clinical applications, e.g., tumor definition[6], image segmentation[7,8], motion estimation[9], evaluation of tumor response or treatment[10], image-guided surgery[11], and adaptive radiotherapy[12]. While DIR can be used to register images from different patients or modalities, this study focuses on registrations between diagnostic computed tomography (CT) image pairs of the same patient.

Despite its wide array of applications, DIR is currently underutilized in the clinic. This is largely due to difficulty in the estimation of DIR accuracy[1,13]. Verification of DIR accuracy is non-trivial and strongly case-specific, as often multiple solutions to the matching process can be found[14]. Common methods to verify DIR accuracy include target registration error (TRE) calculated from anatomical landmarks, contour overlap metrics, intensity-based similarity coefficients between the moving and target images, digital phantoms, and more[4,14-16]. All these DIR verification approaches have drawbacks that have prevented any of them from becoming an acceptable standard. Image similarity metrics computed between the deformable source image and the target image are convenient, but are not directly related to anatomy, and therefore can often not fully describe sub-optimal registrations[17]. Digital phantoms can be useful in ideal scenarios but may not sufficiently capture patient-specific anatomy, image quality, artifacts, and DVF complexity found in clinical cases[18]. Contour-based metrics rely on contour accuracy and may not properly measure accuracy within the contours[14,19]. TRE estimates calculated using anatomical landmark pairs are intuitive and accurate, but defining sufficient landmark pairs is time-consuming and subject to inter-observer variations.

For clinical implementation, DIR would ideally be applied and verified on a patient-specific basis. Due to the aforementioned challenges, however, there exists no standardized approach for such a task[1]. To achieve this goal, preclinical validation of algorithm accuracy is an important first step. This has been demonstrated by the improvement in lung registration over the past few decades. Lung registration is currently one of the most tractable registration scenarios, with mean TREs of 0.7mm to 2mm[20,21]. This is in large part due to existing benchmark datasets in the lung that have supported algorithm development. Of note are the DIRLAB 4DCT and POPI datasets, which have been widely used to verify thoracic registration algorithms in recent years[22-26]. These datasets contain between 75-300 manually labeled blood vessel bifurcation landmarks on thoracic CT scans from the same patient. In addition, Criscuolo et al recently published a publicly available dataset of 30 thoracic CT image pairs with an average of 1262 landmarks per case for algorithm validation[27]. By comparing the initial and deformed locations of the bifurcation landmarks in these datasets, researchers can calculate the TRE of their algorithms and have anatomically focused ground truths that they can reference. Due to the difficulty in identifying landmark pairs, similar datasets in other anatomical sites are currently very limited, and those that exist often do not have sufficient landmarks for robust algorithm validation. One exception is the recent dataset by Zhang et al, which comprised 2220 bifurcation landmarks across 40 liver CT scan pairs[28].

In contrast to the lung, abdominal CTs are one of the most challenging registration scenarios. Complex deformations in the abdomen, inconsistent image content due to the filling of gastrointestinal (GI) organs, image contrast variations, and differences in patient positioning or respiratory status during imaging all make abdominal DIR extremely difficult[29]. This is particularly the case for scans acquired on different days, as these differences evolve over time. Several groups have demonstrated inter-fractional motions of up to 40mm for GI organs[30-35]. Given this additional challenge, we will focus on registrations of this type in this work rather than 4DCT or intra-fractional data.

There are limited options to verify DIR algorithms for abdominal cases because there are no existing ground truth landmark pair datasets by which TRE can be calculated. Other metrics have been previously measured for DIRs on clinical abdominal images. For example, Saleh et al introduced the distance discordance metric and reported values up to 15 mm for the bladder and rectum[36]. Additionally, the mean surface distance estimated on clinical CT images by Xu et al. was approximately 9 to 38 mm for abdominal organs[29]. Large DIR errors and significant uncertainties in these verification metrics highlight the need for DIR validation datasets that can be used to calculate TREs while capturing the high degree of variability found in clinical abdominal images.

Finding sufficient anatomical landmarks in abdominal CT scans for DIR validation is difficult due to a lack of stable high-contrast features in the abdomen. However, iodine-based intravenous (IV) contrast is often administered to patients for abdominal CT scans for the opacification of vascular structures and solid abdominal and pelvic organs[37]. In most cases, IV contrast is recommended for abdominal imaging to support diagnosis or treatment[38]. Blood vessel bifurcations in these contrast-enhanced CT (CECT) images are well-suited for DIR validation because they are abundant, repeatable across different CT scans, and visible with high contrast. These vessels and their bifurcations can reasonably infer complex deformations of nearby tissues because they are flexible and comply with organ deformations. However, automatic vessel segmentation techniques are generally ineffective in the abdomen, making bifurcation identification difficult and laborious. This has precluded their use for abdominal DIR validation, despite the extensive use of such vessel bifurcations in DIR validation for lungs[22,25,26,39].

Automated landmark detection is appealing for DIR validation as it mitigates inter-observer variability in defining landmark positions. There have been several efforts to automate the landmark detection process in the body. A dense set of landmarks was detected in CT and MRI images using scale-invariant feature transform (SIFT) features and matched using an inverse consistent guided technique by Yang et al [40]. This technique effectively identified a high number of landmark pairs, but the landmarks are based on image features and not anatomy and therefore can be difficult to verify. To overcome this limitation, Fu et al.[41] detected landmarks on the probability maps of lung vasculature, established correspondence with an approximate registration, and refined spatial position with a pseudo-Siamese deep learning network. Cazoulat et al[42] utilized a similar approach, establishing vessel bifurcation landmark pairs in lung CT images by registering their segmented vessel maps. The recent dataset published by Criscuolo et al. in the lung also incorporated these methods, establishing blood vessel bifurcation pairs with vessel segmentation and deformable registration[39]. Similar work was utilized by Zhang et al to generate a liver DIR validation dataset, including multiple registration methods for landmark matching to reduce algorithm bias[43].

With this in mind, we develop and apply a semi-automatic pathway that can efficiently identify a large number of vessel bifurcation landmark pairs in abdominal CECT scans. This workflow expands upon the dataset generation methodology from Zhang et al[28] and Criscuolo et al[27] by adding several key steps; deep learning segmentation was used to overwrite inconsistent abdominal organ intensities, multiple local registrations per image were used to match bifurcation pairs, and an iterative bifurcation alignment technique was employed after landmark verification. These steps help to mitigate bias from the registration algorithm used to match the landmarks, a key limitation of the datasets from Zhang and Criscuolo et al, as well as increase the positional accuracy of the final pairs. We applied this workflow to 30 abdominal CT scan pairs acquired on different days to develop a first-of-its-kind, publicly available DIR validation dataset that can be used to support abdominal DIR development.

## 2 Acquisition and validation methods

### 2.1 Data sources

30 pairs of high-quality CECT images were included in this study. A scan pair refers to two scans acquired from the same patient at different times, as a part of treatment or another clinical workflow. Of these 30 image pairs, two were from the Anti PD-1 Immunotherapy Lung (PD-1 Lung) dataset[44,45], 18 were from the Anti-PD-1 Immunotherapy Melanoma dataset (PD-1-Melanoma)[46], and 10 were acquired from Barnes Jewish Hospital (BJH) with IRB approval. Protected health information (PHI) was removed from the BJH scans and their DICOM headers.

Of these 30 scans, 25 are also included in our liver dataset[28]. Using the landmark pairs from both the liver and abdominal vessels can offer even more comprehensive information for algorithm validation. Table 1 highlights which cases are covered in both datasets, while our GitHub describes what case indices in this dataset correspond with the cases in the liver dataset.

Scans in the PD-1 Melanoma dataset were acquired from patients receiving anti-PD1 immunotherapy for Melanoma treatment in 2016. For each patient, one CT was acquired pre-treatment and the second was acquired at a follow-up timepoint. Similarly, scan pairs from the PD-1 Lung dataset were from lung cases treated with anti-PD-1 Immunotherapy in 2016, with each case containing one pre-treatment and one follow-up image. The BJH data encompasses a variety of patient and scan conditions. The patient in each scan received both oral and intravenous contrast. For a full description of the scan parameters used to acquire the images, please reference the supplemental information document or the DICOM headers posted on our GitHub at https://github.com/deshanyang. The source dataset, case identifier, time between scans, and resolution of each image in the dataset can be found in Table 1 in Section 3.

### 2.2 DIR validation and landmark pair identification pipeline

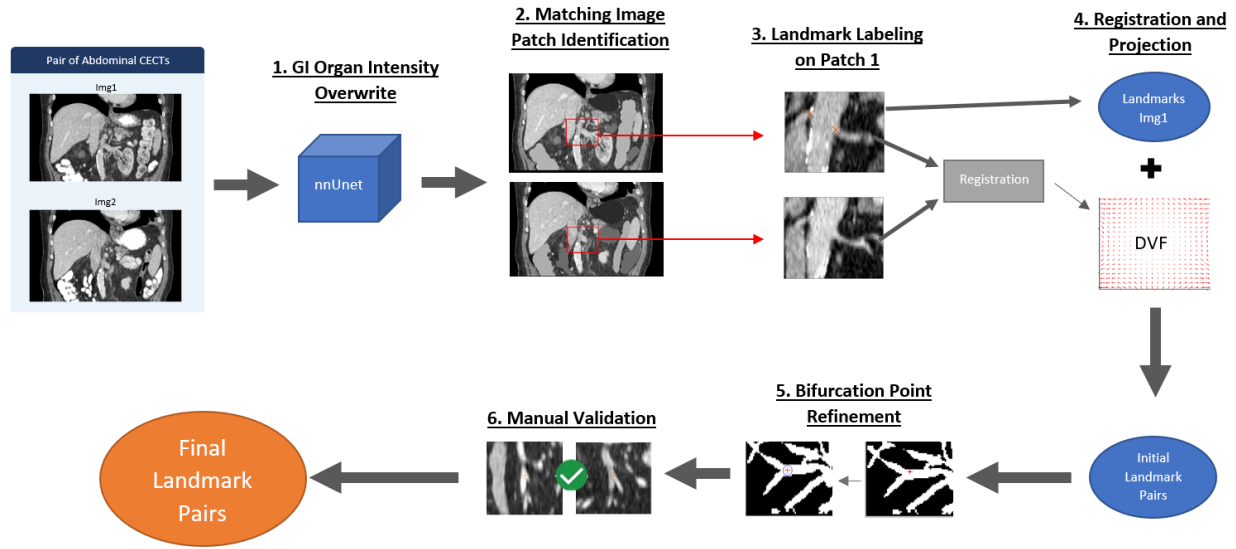

*Figure 1: Landmark pair identification pathway. This pathway was applied in its entirety to each of the 30 image pair cases in this dataset. Steps 1-4 are described briefly in the following section, with a more extensive discussion of steps 5 and 6, as these are the limiting factors in the accuracy of the landmark pairs. For further detail on steps 1-4, please reference the supplemental document*

The landmark pair identification workflow used in this study is illustrated in Figure 1. Given an image pair, the stomach, small intestine, duodenum, and colon were segmented in both images using a deep-learning model[47,48], and segmentation errors were manually corrected. The image intensities of these organs were overwritten to organ-specific constant values to help guide registration in later steps. Each image pair was deconstructed to a set of smaller matching image patches, defined in both images by manually identifying matching structures, i.e. corresponding large vessels or organ boundaries. Blood vessel bifurcation landmarks were manually labeled on the first image in each pair of image patches. The patch pairs were then individually registered and the resultant DVFs were used to project the landmarks labeled in the first image into the second image. The projected landmark positions were manually verified to form the initial landmark pairs. For more details on steps 1-4, please reference the supplemental document. In the following section, we describe how we further refined these initial landmark positions (step 5) to ensure precision and avoid bias in their use in DIR validation.

## 2.3 Bifurcation Point Refinement

Steps 1-4 of the semi-automatic landmarking pathway shown in Figure 1 identify an initial set of landmark pairs that can be used for DIR validation. However, these steps are limited by their reliance on a single registration method, pTVreg, to project labeled landmarks from Image 1 to Image 2.

Our goal was that the landmark pairs in this dataset could serve as a verification method for algorithms beyond the accuracy level of current methods. As previously mentioned, overwriting the abdominal organ intensity and selecting guiding image patches were intended to increase projection accuracy. In addition, the manual verification step ensured those landmarks improperly projected were excluded from the final dataset. However, the final landmark pairs and accuracy metric calculated using them may still be biased to the DIR method, pTVreg, used in the landmark projection step, or similar algorithms.

To eliminate this potential bias, we employed two approaches to refine the landmarks to the optimal vessel bifurcation spots, depending on their bifurcation type. We defined two different types of vessel bifurcations based on vessel diameter. Type 1 bifurcations were those comprised of vessels of similar diameters, while type 2 bifurcations were those where the vessels had dramatically different diameters, as is often seen in the abdomen. This is shown in Figure 2.

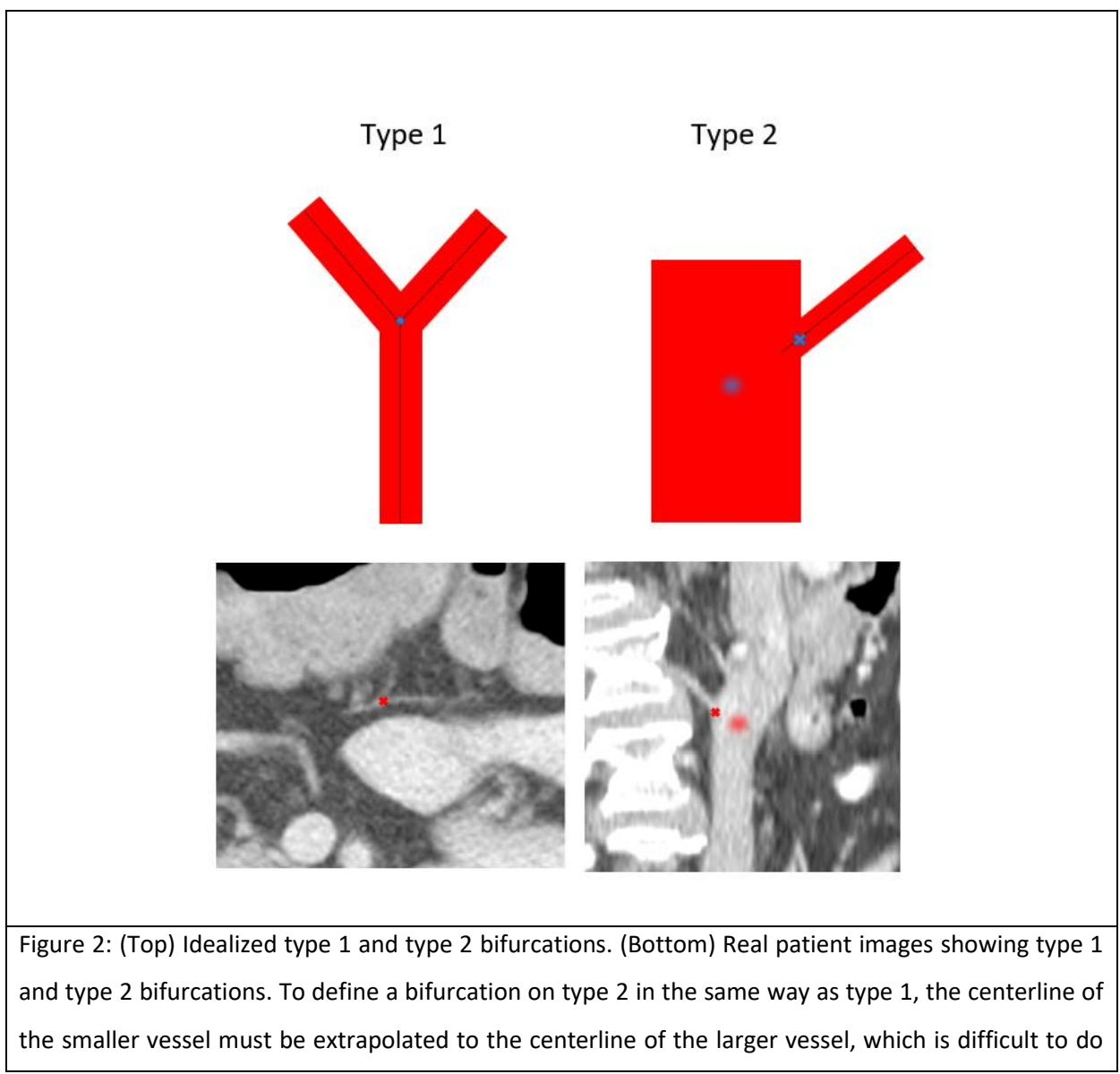

Figure 2: (Top) Idealized type 1 and type 2 bifurcations. (Bottom) Real patient images showing type 1 and type 2 bifurcations. To define a bifurcation on type 2 in the same way as type 1, the centerline of the smaller vessel must be extrapolated to the centerline of the larger vessel, which is difficult to do

> with precision, as shown in the blurred circles. The curvature of the vessels and the image artifacts could further increase bifurcation uncertainty. It is less ambiguous for type 2 bifurcations to place the landmark at the intersection point (shown by x)

The cutoff point at which vessels were considered type 2 was when the larger vessel comprising a bifurcation was ≥2.5 times larger in diameter than the smaller one. The definitions of type 1 and type 2 landmarks can also be used to refer to a landmark pair, as opposed to an individual bifurcation. This is because the bifurcation should occur on the same anatomy in both images, and the vessel diameter should be consistent between images 1 and 2. Moving forward, landmark pairs can also be referred to as type 1 or 2.

Type 2 landmark pairs in our dataset were refined manually, while type 1 bifurcations were refined using an iterative sphere growing approach described in the following section.

### 2.3.1 Sphere Growing Algorithm for Type 1 Bifurcations

For type 1 vessels, segmenting and identifying the centerline is one way to label bifurcations, but automated vessel segmentation in abdominal CTs is challenging and unreliable. Therefore, we introduced a novel method using an iteratively grown sphere to define bifurcation points, improving spatial correspondence between bifurcation landmarks, and permitting validation of high-precision algorithms. This method also reduced bias from the registration used in step 4 of Figure 1. Finally, we hoped to develop a robust method of defining a bifurcation point, which was a poorly defined problem.

The sphere-growing process, inspired by Li et al.'s minimal path strategy[49], started from the bifurcation point labeled by the researcher or projected by pTVreg. A 100mm³ sub-volume centered on the bifurcation was upsampled to ~0.7mm isotropic resolution. Gradients of the sub-volume were computed and convolved with a low-pass filter (σ=1 pixel). A sphere with a 0.5-voxel radius was initialized at the bifurcation and iteratively expanded, with radius growth balanced by a constant outward internal force and gradient-based opposing force, ensuring it fits within the vessel's border. The sphere center was adjusted based on gradients to fit the sphere within the vessel's widest portion, aligning with the bifurcation. For our dataset, 30 iterations were sufficient for the algorithm to converge.

Equations 1-3 model this process in an image volume $I$. The sphere center, $\vec{c}$, and radius, r, were iteratively updated in equations 2 and 3 using the internal force $f_{int}$ and opposing force $\vec{\mu}$, directing the center to the vessel's widest point but not growing beyond the vessel border. $\lambda_1$ and $\lambda_2$ were scaling factors, set respectively to 0.2 and 0.3 in our implementation, while $\vec{X}$ is the current voxel index and S gives all the indices of the image volume I.

$$\vec{u} = -(1-I)\nabla I/|\nabla I| \qquad (1)$$

$$\vec{c}_{n+1} = \vec{c}_n + \lambda_1 \overline{\sum_{\vec{X} \in S} \vec{u}(\vec{X})} \qquad (2)$$

$$r_{n+1} = r_n + f_{int} + \lambda_2 \overline{\sum_{\vec{X} \in S}(\vec{X}-\vec{c}) \cdot \vec{u}(\vec{X})} \qquad (3)$$

For this dataset, sphere growing was applied to all type 1 landmark pairs with images thresholded between -160 and 240 HU to enhance vessel contrast. The sphere's iterative position in each case was monitored, and if growth diverged due to vessel shape or overlapping anatomy, region growing on a vesselness[50] image generated a mask for rerunning the algorithm. For unresolved cases, manual vessel segmentation was used. Of the refined landmark pairs, 68% were grown on the original image, 24% on the automatic vessel mask, and 8% on the manual vessel segmentation mask.

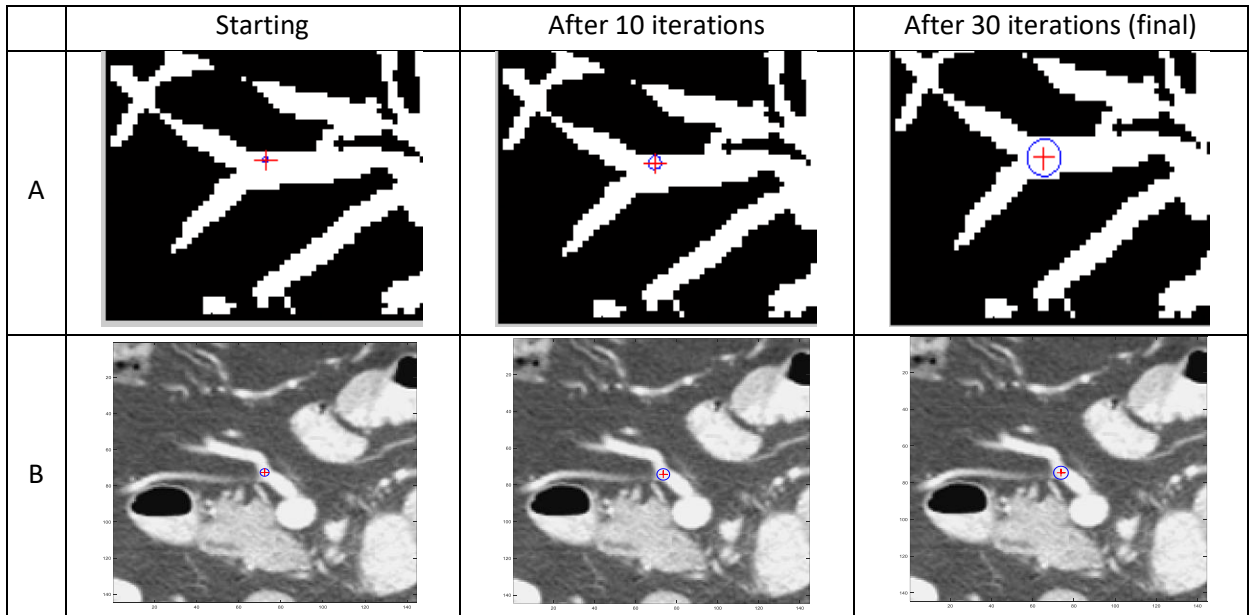

*Figure 2: Demonstration of the sphere-growing method to find the bifurcation centers. a) The iterative process on a slice of a pre-segmented vessel mask. The center of the sphere moves towards the widest point of the vessel as the sphere grows, which should be where the bifurcation occurs. b) The iterative process directly on a patient image. Note that the mask and the image are not of the same vessel*

### 2.3.2 Manual Placement of Type 2 Bifurcations

The bifurcation point for type 2 bifurcations was defined as the point at which the centerline of the smaller vessel intersects the surface of the larger vessel. Because there was less ambiguity in the placement of this point than in type 1 landmarks, landmark pairs of this type were manually defined by a researcher on images 1 and 2. To assess the accuracy of this manual process, two independent observers performed the same task on a subset of the final landmarks, and the variability was calculated. This is further described in section 2.4.

### 2.3.3 Manual Validation of the Final Landmark Positions

After the refinement of the bifurcation positions, the final landmark pair positions were validated by two independent observers. The observers verified that each bifurcation pair occurs on a true bifurcation, that the bifurcation point was clearly defined, and that the locations of the placement on image 1 and image 2 were accurate. If any of these were deemed an issue, the landmark pair was flagged, and a final check by the primary researcher was done to decide whether to remove the bifurcation or manually adjust its final positioning.

By manually adjusting some of these final landmark positions, we reintroduced inter-observer variability to type 1 landmarks that was amended with our sphere growing algorithm. These landmark positions that were manually adjusted were included in a separate version of the data, available at the same repository labeled as "flagged landmarks".

## 2.4 Landmark Pair Assessment
### 2.4.1 Inter-Observer Variability of Manually Placed Landmarks

Out of the final 1895 landmark pairs in this dataset, 510 were comprised of type 2 landmarks. These landmark pairs were labeled manually according to the definition described in 2.3, and therefore subject to inter-observer variability. To use the landmark pairs for DIR evaluation, it was critical to ensure a landmark placed in Image 2 corresponded accurately with the matching landmark labeled in Image 1.

To estimate the inter-observer manual landmarking variability, 50 random type 2 bifurcation cases were selected. A pair of small 3D image patches of 100 mm in each dimension were extracted from both images centered at the landmark pair. The image patch pair was provided to two independent observers. The center of image patch 2 was slightly offset by randomly shifting the patch ≤3 slices off the center in each axis, and the landmark position on that image patch remained unlabeled. The landmark position on image patch 1, as labeled by the primary researcher, remained visible. The primary researcher and two independent observers must use the position labeled on image patch 1 to identify the corresponding position on image patch 2 to the best of their abilities. The landmark error was calculated as the average three-dimensional Euclidean distance between the position labeled by the primary researcher and each independent observer. This error averaged 1.2mm +/- 0.7mm. We therefore inferred that our manually refined type 2 landmarks were beyond current abdominal DIR accuracy and were no longer biased towards pTVreg or similar algorithms.

### 2.4.2 Landmark Positions Refined Using Sphere Growing

The sphere growing method used in this work was meant to standardize the way bifurcations were defined, remove the bias introduced by the registration algorithm used to match the landmarks, and increase the precision of anatomical correspondence of the landmark pairs. Two more phantom studies were designed and performed to assess if the procedure was effective in these tasks.

The first study assessed if the algorithm was more accurate in matching the landmark position in image 2 than manual landmark labeling. 50 random landmarks across all 30 cases were chosen. A small image patch, 200 mm in each dimension, centered on each landmark, was extracted and upsampled to an approximate isotropic voxel size of 0.7 mm. This image patch was then put through a series of transformations to form a second patch simulating a second clinical image, but where the ground truth deformation was known. The transformations consisted of a random rotation through an angle of 0-50 degrees about a random axis of rotation, an image up or down scaling within 10%, and an artificial sinusoidal deformation field. Significant amounts of deformation chosen for this phantom synthesis task were aimed to replicate the variations found in the orientation and structure of abdominal vessels.

The sphere growing algorithm described in 2.3.1 was then performed on both the image patch 1 and the artificial phantom image patch 2, resulting in two landmark positions. The transformation matrices generated in the phantom creation process were used to determine the ground truth position in image patch 2 for the sphere-grown position in image patch 1. This ground truth position was compared to the sphere-grown position on image patch 2 to determine the error. In this way, the precision of the sphere growing algorithm was estimated using digitally deformed image patch pairs.

The phantom image patch pairs were also provided to two independent observers in the same manner as the inter-observer study on type 2 landmarks. The landmark position on the phantom image patch 2 was hidden and the observers were tasked with manually placing the landmark where it should match its position on the image patch 1. The distances from the manually placed landmark in image patch 2 to the ground truth position calculated by the transformation matrices give the manual landmarking error.

By this procedure, across 50 phantom pairs, the manual labeling error by two observers averaged 1.3mm +/- 0.75mm. In comparison, the sphere growing error was 0.7mm +/- 0.7mm on average. Running a paired t-test on this data, we get a strongly significant p value of 1.7e-6. The sphere growing method was thus confirmed to be significantly more precise than manual landmarking.

The second study was designed to confirm that the sphere growing procedure contributed additional landmark pair accuracy over just using pTVReg. A similar digital phantom generation procedure as above was applied to 100 randomly selected landmarks. However, instead of having an observer place the landmark position in image 2, pTVreg was used to register the phantom image pair and project the landmark position from image 1. The ground truth transformation information was used to calculate the accuracy of the projected landmark position. The registration error of PTV was 2.7mm +/- 6.6mm, while the sphere growing error was 0.7mm +/- 0.6mm. A paired t-test on these 100 observations has a p value of 0.0035. It was therefore confirmed that our sphere growing procedure was more precise than strictly using pTVreg to establish landmark correspondence. It is worth noting, however, that by excluding the 14% of landmarks with an error greater than 5mm, the PTV accuracy converged to 0.45mm +/- 0.66mm. While this improved accuracy suggested that pTVReg may be effective if outliers were identified, relying solely on pTVReg projections introduced biases inherent to the algorithm. In contrast, the sphere-growing method avoided such biases and demonstrated higher precision than manual landmarking. Although limitations existed in the phantom generation process, the findings supported the robustness and reliability of the sphere-growing method for defining landmark positions.

# 3 Data Format and Usage Notes

The final landmark data after processing is shown in Table 1.

Table 1: The list of image pairs and detected landmarks. [1]Melanoma = Anti-PD-1-Melanoma. [2]Lung = Anti-PD-1-Lung. [3]BJH = Barne Jewish Hospital. The image cases in italics are also included in the liver vessel bifurcation dataset from Zhang et al[28]. The corresponding case indexes in the liver dataset can be found on our Github.

| # | Num of Landmark Pairs (Normal, Flagged) | Source | Case # in Source | CT Scan Interval (days) | Image 1 | | Image 2 | |
|---|---|---|---|---|---|---|---|---|
| | | | | | Slice Thickness (mm) | In-plane resolution (mm) | Slice Thickness (mm) | In-plane resolution (mm) |
| 1 | 57 (52,5) | [1]Melanoma | 03 | 111 | 2.5 | 0.82 | 2.5 | 0.82 |
| 2 | 60 (58,2) | Melanoma | 06 | 84 | 2.5 | 0.82 | 2.5 | 0.82 |
| 3 | 57 (52,5) | Melanoma | 11 | 104 | 2.5 | 0.82 | 2.5 | 0.82 |
| 4 | 57 (51,6) | Melanoma | 12 | 97 | 2.5 | 0.70 | 2.5 | 0.70 |
| 5 | 79 (73,6) | Melanoma | 13 | 98 | 2.5 | 0.78 | 2.5 | 0.78 |
| 6 | 54 (47,7) | Melanoma | 14 | 98 | 1.25 | 0.70 | 2.5 | 0.70 |
| 7 | 60 (56,4) | Melanoma | 15 | 115 | 2.5 | 0.78 | 2.5 | 0.78 |
| 8 | 69 (64, 5) | Melanoma | 19 | 111 | 2.5 | 0.78 | 2.5 | 0.78 |
| 9 | 122 (114,8) | Melanoma | 20 | 99 | 2.5 | 0.82 | 2.5 | 0.82 |
| 10 | 43 (42,1) | Melanoma | 22 | 37 | 2.5 | 0.74 | 2.5 | 0.74 |
| 11 | 64 (60,4) | Melanoma | 25 | 42 | 2.5 | 0.86 | 2.5 | 0.86 |
| 12 | 85 (82,3) | Melanoma | 27 | 104 | 2.5 | 0.86 | 2.5 | 0.86 |
| 13 | 56 (50,6) | Melanoma | 40 | 83 | 2.5 | 0.94 | 2.5 | 0.94 |
| 14 | 56 (53,3) | Melanoma | 47 | 50 | 2.5 | 0.7 | 2.5 | 0.7 |
| 15 | 117 (114,3) | [2]Lung | 03 | 76 | 2.5 | 0.78 | 2.5 | 0.78 |
| 16 | 47 (44,3) | [2]Lung | 11 | 40 | 2.5 | 0.78 | 2.5 | 0.78 |
| 17 | 78 (72,6) | Melanoma | 39 | 97 | 2.5 | 0.86 | 2.5 | 0.86 |

| 18 | 90 (86,4) | Melanoma | 37 | 89 | 2.5 | 0.90 | 2.5 | 0.90 |
| 19 | 79 (77,2) | Melanoma | 33 | 98 | 2.5 | 0.70 | 2.5 | 0.70 |
| 20 | 64 (62,2) | Melanoma | 31 | 198 | 2.5 | 0.86 | 2.5 | 0.86 |
| 21 | 65 (65,0) | [3]BJH | - | 1831 | 1.00 | 0.98 | 2.00 | 0.87 |
| 22 | 93 (90,3) | BJH | - | 97 | 3.00 | 0.87 | 3.00 | 0.96 |
| 23 | 60 (58,2) | BJH | - | 278 | 2.00 | 0.69 | 3.00 | 0.89 |
| 24 | 30 (30,0) | BJH | - | 92 | 2.00 | 0.68 | 2.00 | 0.66 |
| 25 | 45 (43,2) | BJH | - | 564 | 3.00 | 0.84 | 3.00 | 0.89 |
| 26 | 59 (56,3) | BJH | - | 143 | 2.00 | 0.94 | 2.00 | 0.89 |
| 27 | 62 (60,2) | BJH | - | 2198 | 2.00 | 0.75 | 2.00 | 0.73 |
| 28 | 78 (74,4) | BJH | - | 168 | 2.00 | 0.66 | 2.00 | 0.66 |
| 29 | 72 (71,1) | BJH | - | 2324 | 2.00 | 0.79 | 2.00 | 0.74 |
| 30 | 44 (39,5) | BJH | - | 506 | 3.00 | 0.78 | 3.00 | 0.76 |

### 3.1 Dataset Overview

The data was stored in the Zenodo online data repository at https://doi.org/10.5281/zenodo.14362785. Instructions for getting started with the dataset can be found on our GitHub at https://github.com/deshanyang/Abdominal-DIR-QA. Image and landmark data were saved both as NIfTI files and MATLAB .mat files

27%, or 507 of the final 1895 landmark pairs, were type 2 bifurcations that were placed manually. The remaining 73%, or 1388 landmark pairs, were refined using the sphere growing algorithm.

Voxel sizes and intensities of the images were saved in the format of the original scans. This gives researchers flexibility in how they process the data. Additional information on scan parameters of images in the dataset, including in-plane resolution, slice thickness, kV, and exposure, was provided in the supplemental document.

### 3.2 Use

A guide for using the landmarks can be found on our Github at https://github.com/deshanyang/Abdominal-DIR-QA. Researchers may visualize the landmark pairs in the dataset using MatchGui, a MATAB-based tool we developed in-house. Code and use instructions for MatchGui can also be found on our Github page. The two images in each case can be registered using any DIR method and the corresponding landmarks of the case can be used to verify the DIR results and compute TREs.

The images included in the dataset came from a variety of CT scanners, and had a range of image quality and voxel sizes. However, to maximize the number of landmark pairs, the images were of mostly high quality. DIR accuracy estimation in poor image quality cases is of concern for robust validation. Therefore, we recommend that images of worse quality be simulated using noise addition and image transformations. In this way, challenging DIR situations can be simulated and robustly tested using the provided landmark pairs.

## 4 Discussion

We successfully developed a first-of-its-kind DIR validation dataset for abdominal CT scans in this work. We identified 1895 landmarks across 30 image pairs, with 100 additional unrefined landmarks included for completion. The estimated landmark TREs from phantom studies were 0.7-1.2 mm, well beyond reported registration errors for abdominal CT scenarios, which exceeded 10 mm[29]. This dataset could be useful in verifying new DIR methods and establishing baselines for current methodologies. In addition, 25 of the scans included in this dataset were also included in our previously published liver DIR landmark dataset, which contained labeled blood vessel bifurcation pairs within the liver. Using these datasets in tandem will give researchers comprehensive information on the performance of their algorithms within and outside of organs.

Beyond the dataset itself, the workflow (Figure 1) used to generate the landmark pairs can be reasonably automated with further efforts and employed for patient-specific DIR validation. For example, Step 1 involved segmenting the abdominal organs with deep learning and manually correcting any mistakes. Medical image segmentation accuracy using deep learning methods has increased significantly in recent years, and the number of papers published on deep learning medical image processing papers on PubMed has increased from ~2000 to ~16,000 from 2017 to 2021[51]. We believe the manual correction step can be increasingly limited or removed entirely given improved deep-learning segmentation methods in abdominal CTs. Bifurcation labeling, Step 3 in Figure 1, was performed manually in this study. However, automated landmark labeling is an established research field and methodologies like corner detection or improved blood vessel segmentation can be used for automatic detection. Finally, the validation of landmark pairs, Step 6 in Figure 1, can be automated with deep learning image-matching approaches.

Our sphere-growing algorithm was robust to significant anatomical distortions which would make manual delineation difficult. Our phantom study confirmed this by estimating the manually placed landmark pair uncertainty as 1.3mm +/- 0.75mm and 0.7 mm +/- 0.7 mm for the sphere-grown pairs. However, this improvement was reliant on the sphere-growing algorithm being converging correctly. In cases with overlapping vessels or poor contrast, the vessel segment mask was often required for the sphere-growing algorithm to compute the bifurcation center properly. More sophisticated versions of the sphere-growing method that can include anatomical information beyond image intensity may be more robust to issues like this.

We identified inter-observer variability of manually placed landmark pairs as a key limitation of prior datasets. Despite this, 510, or 27% of the final landmark pairs in this dataset were type 2 landmarks, which we chose to place manually. The estimated inter-observer variability of this placement was 1.2mm +/- 0.7mm. Developing an automated procedure similar to the sphere growing procedure applied to type 1 landmarks could help reduce this variability. However, vessel segmentation and skeletonization algorithms generally perform poorly for vessels with largely different diameters, and there are few alternatives. Given this, and the fact that manually labeled bifurcations are not subject to bias from the projection step, it is the authors' opinion that manual labeling was the best approach for type 2 landmark pairs. However, future iterations of the dataset would benefit from automating the placement of these pairs as well.

# 5   Conclusions

This work represents a first-of-its-kind abdominal CT landmark pair dataset that can be used for DIR validation and algorithm development. The data comprises a variety of patient and scan conditions, permitting comprehensive validation across clinical scenarios. This dataset is publicly available and could serve as a baseline reference for researchers verifying registration algorithms in abdominal CTs.

# Acknowledgments

This research was supported by the National Institute of Biomedical Imaging and Bioengineering (NIBIB) grant R01-EB029431. The results here are in whole or part based upon data generated by the TCGA Research Network.[45]